# A Diversified Multi-Start Algorithm for Unconstrained Binary Quadratic Problems Leveraging the Graphics Processor Unit

Mark Lewis
Missouri Western State University, Saint Joseph, Missouri 64507, USA
mlewis14@missouriwestern.edu

**Abstract**

Multi-start algorithms are a common and effective tool for metaheuristic searches. In this paper we amplify multi-start capabilities by employing the parallel processing power of the graphics processer unit (GPU) to quickly generate a diverse starting set of solutions for the Unconstrained Binary Quadratic Optimization Problem which are evaluated and used to implement screening methods to select solutions for further optimization. This method is implemented as an initial high quality solution generation phase prior to a secondary steepest ascent search and a comparison of results to best known approaches on benchmark unconstrained binary quadratic problems demonstrates that GPU-enabled diversified multi-start with screening quickly yields very good results.

**Introduction**

Concepts associated with scatter search outlined in (Glover, 1998) describe creating and evaluating a large number of diverse solutions. However, when the objective function is complex, numerous evaluations of the objective function adds considerably to the time complexity of a search algorithm. An approach to reduce the the time complexity of a combinatoric solution space is to implement a screening technique to search only those neighborhoods expected to yield high quality solutions.

A significant improvement has recently been made avaialable to address both these topics. The Compute Unified Device Architecture (CUDA) is a parallel computing platform and programming model for harnessing the power of the Graphics Processor Unit (GPU). This parallel computing power can dramatically decrease the time needed for objective function evaluation and is particularly beneficial as problem size increases. A benefit of being able to quickly evaluate large solution sample spaces is the ability to quickly calculate effective objective function value screening levels.

This paper investigates offloading the evaluation of diverse solutions from the host Central Processing Unit (CPU) to the GPU followed by a parametric screening technique and steepest ascent search. This approach is applied to the optimization of the Unconstrained Binary Quadratic Optimization Problem (UBQP). The UBQP has been studied for several decades and has more recently attracted attention as an effective modeling framework for a wide variety of both linear and quadratic problems.

The UBQP takes the form:

$$(P) \text{ Maximize } x_o = f(x) = x^t Q x \quad x \in X$$

where $Q$ is an *n*-by-*n* square symmetric matrix of real or integer coefficients and $x_i \in \{0, 1\}$. The UBQP is often referred to by its objective function $xQx$.

Although there are very fast and effective methods to evaluate 1-bit flips in UBQP, there is no straight-forward approach to calculating xQx when making multiple changes. straight-forward implementation to evaluate the UBQP's quadratic objective function is $O(n^3)$. Offloading the evaluation to the GPU provides enormous benefit. Table 1 illustrates respective times spent by a CPU and a GPU when evaluating 1000 random solutions to UBQP of increasing size. The calculation of xQx by the GPU is marginally affected by increases in problem size, hence the motivation for this study.

|        | Time (sec) |     |
|--------|------------|-----|
| Q size | GPU        | CPU |
| 2500   | 0.7        | 46  |
| 5000   | 2.0        | 235 |
| 6000   | 3.0        | 357 |
| 7000   | 3.5        | 512 |

Table 1. GPU speed improvements over CPU for calculating 1000 random UBQP

The first part of the paper presents a brief literature survey of scatter search and the UBQP. The second part presents the general problem description of screening diversified multi-start solutions for subsequent solution polishing. The third part presents an algorithmic implementation and computational results.

**Literature Survey**

Early work with UBQP by ( Barahona, et al., 1988) dealt with the physics problem of finding ground states of spin glasses with exterior magnetic fields. Subsequently, published UBQP applications include max-cut (Boros & Hammer, 1991) and (Kochenberger, et al., 2013), machine scheduling (Alidaee, et al., 1994), maximum clique (Pardalos & Xue, 1994), number partitioning (Alidaee, et al., 2005) and max 2-sat (Kochenberger, et al., 2005).

The GPU is a multi-processor that specializes in manipulated large blocks of data in parallel using, for example, 2880 cores. In contrast the CPU specializes in general purpose implementations of natural language pseudo-code allowing diverse logic flows with, for example, 4 cores. The CUDA platform was made available by NVIDIA in about 2007 to allow software developers access to the power of their GPUs and now many high performance computers utilize a GPU architecture.

**Pseudocode**

The basic approach is to generate a large number of diverse solutions and those that pass a value determined by a screening function $T(\lambda) = \text{Mean} + \lambda(\text{Max} - \text{Mean})$. Solutions passing the screening are selected for further improvement via a steepest ascent approach based on 1-bit flips that improve the objective. The screening function $T(\lambda)$ generates a value based on the mean xQx value over all samples and a parametric percentage of the difference between the current max and the mean. Thus $T(\lambda)$ will change as the mean and max change and with changes in $\lambda$. After a local optima is discovered and there are no beneficial single bit flip changes, then diversifications are applied until a solution passes the screening value.

The above concepts were implemented in CUDA C according to the pseudocode described below in Figure 1. which employs concepts from (Glover, 1998), specifically the Diversification Generation schemes in the Initial Phase prior to the Scatter Search / Path Relinking Phase.

The approach implemented leverages the parallel compute capabilities of the GPU in two important ways. First, when computing the expected value of xQx using a large (e.g. 1000 samples) number of randomly generated x, we use the GPU, which excels at matrix multiplication. Secondly, the evaluation of xQx during the Diversify and local improvement loop is also faster on the GPU because the Diversify routine often changes a large number of variables. In contrast, the effect on the objective function of flipping one bit in x is very quickly calculated by the host CPU during the steepest ascent improvement phase by using the 1-flip method outlined in (Glover, et al., 2002).

There are several parameters that affect the search and all are automatically generated with default values. Parameters include: the total number of diversify-screen-search iterations to perform; the number of variables to be changed during diversification is based on the loop counter; and how many sample solutions to use for the calculation of the average objective value -- which was set at 1000 for these tests. Additional parameters include setting the screening value T parameters: max, mean and lambda. In the tested implementation the mean is the average xQx value derived during sampling, the max is initially the best_starting value which is the partial first derivative of Q with respect to $x_i = 1$, and lambda is initially set to $\lambda = \text{Max} / \text{Mean} = \text{Starting\_solution} / \text{Mean}$.

Figure 1. General Algorithm Overview showing division of tasks between CPU and GPU

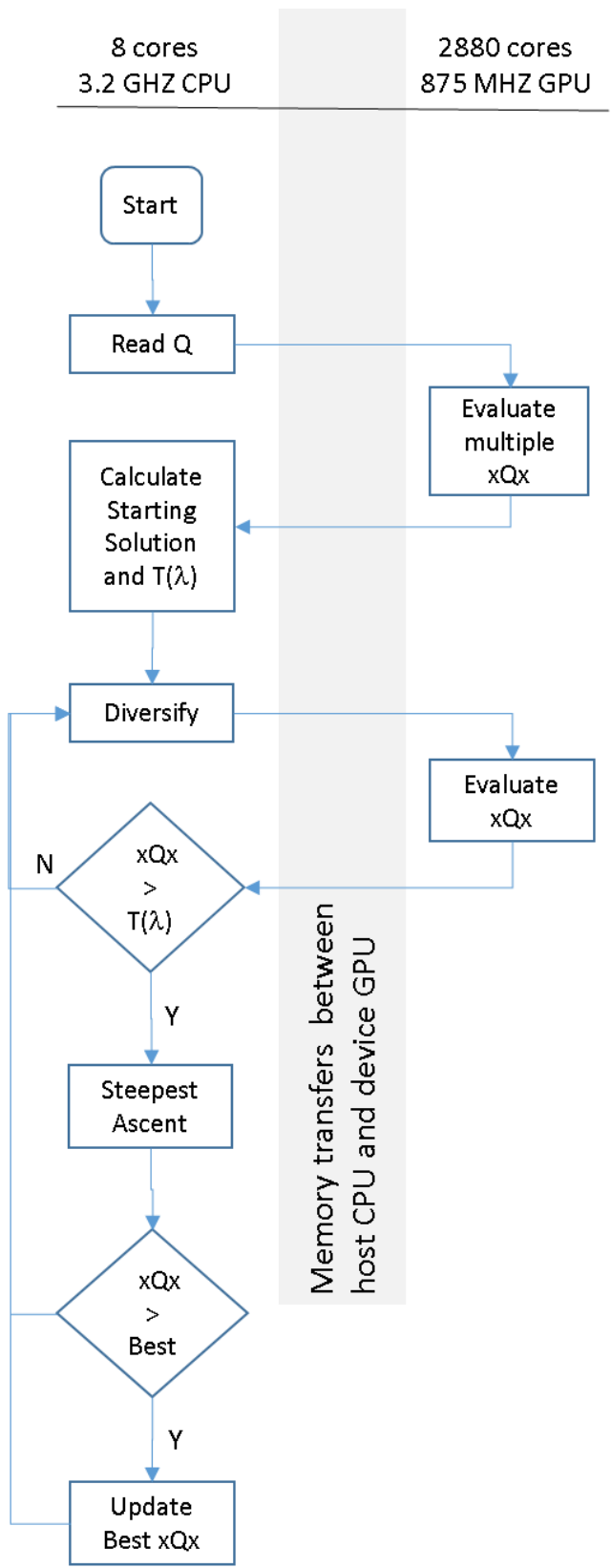

Figure 2. Pseudocode

```
Q_matrix ← ReadQ( Q_file_name );
MoveQtoGPU( Q_matrix );
    Avg_starting_sol_value ← EvaluateRandomStarts ( Q_matrix, Q_cols, num_samples );
Starting_solution ← CalculateFirstDerivativeSolution(Q_matrix, Q_cols);
T ← CalculateT(Avg_starting_sol_value, Starting_solution_value);
CopySolution(x, Starting_solution);

For i = 0 to num_iterations do
    if i > 0 then
         Diversify ( x, best_xQx, i, Q_cols );
    end if
    xQx ← Evaluate( x );
    if xQx > Screening_value then
         PerformSteepestAscent( x, xQx, max_its );
         if xQx > best_xQx_value then
              UpdateBestAndT( best_xQx_value, T, λ , x, best_xQx );
         end if
    end if
End for

Return ( best_xQx_value, best_xQx );
```

The algorithm progresses as follows and as shown in Figure 2. First a Q matrix is read and CPU memory is allocated for x and Q which is then transferred to the GPU. Large memory transfers between the host and device use the slowest memory bus and this Q block of memory is the largest allocation, for example, a 5000 variable problem allocates space for 25 million data elements, typically float or double data types. Luckily, the transfer of Q occurs only once. Our implementation did not take advantage of the CUDA toolkit's (nVidia, 2015) freely available library routines optimized for dense or sparse matrix multiplication. We developed our own CUDA matrix multiplication routine to allow for customization.

In order to compute values for the screening parameters T and lambda, a large number of random solutions are generated and evaluated by the GPU. The CPU is used to calculate the average xQx value and to calculate a starting solution based on the partial first derivative of Q with respect to each $x_i$ evaluated at $x_i = 1$. In other words, simply sum the $i^{th}$ row of Q associated with $x_i$ and if that sum is positive, then set $x_i = 1$ for the starting solution.

After determining T, the main loop is iterated a specified number of times or for a time period. It consists of diversification, evaluation, screening and improvement. Different diversification approaches based on blending (or breeding) two solutions to generate new ones were imlemented. The diversified solution's objective function is evaluated by the GPU and if it exceeds the screening level, then a steepest ascent search begins, terminating when no improvements are possible or a maximum number of flips have been made. No checks for cycling nor tabu lists were implemented. If a new best optimal value is found, then that solution is stored and the screening parameters updated.

The above pseudocode was written and compiled using Visual Studio 2013 with the CUDA runtime environment enabled and then implemented on a host CPU Intel i7-2600 processor running at 3.4 GHz, with 12 GB RAM and an nVidia GTS 780ti GPU with 2880 CUDA cores running at 875 MHz.

The first set of problems tested are the ten 2500 variable sparse ORLIB UBQP. The first set of problems is from the ORLIB (Beasley, 1990) collection and consists of ten sparse instances (linear and quadratic density = 0.1), all of size 2500 variables. Our code performed well on these consistently finding solutions within 0.5% of best reported (Wang, et al., 2012) in an average 3.5 seconds. This research focuses on quickly finding high quality solutions as part of an initial phase of a search, followed by a solution polishing or improvement phase such as path relinking. No one problem stood out as being either much more difficult or easy.

| ORLIB # | best reported solution in literature | MMS (Massive Multi-Start) | % Diff from best solution | Times (sec) MMS time to solution | Wang & Glover time |
|---|---|---|---|---|---|
| 1 | 1515944 | 1513037 | 0.19% | 4.7 | 11 |
| 2 | 1471392 | 1466791 | 0.31% | 2.1 | 101 |
| 3 | 1414192 | 1409367 | 0.34% | 2.1 | 49 |
| 4 | 1507701 | 1504452 | 0.22% | 3.7 | 6 |
| 5 | 1491816 | 1487296 | 0.30% | 3.9 | 14 |
| 6 | 1469162 | 1464082 | 0.35% | 2.2 | 25 |
| 7 | 1479040 | 1474810 | 0.29% | 4.2 | 48 |
| 8 | 1484199 | 1481496 | 0.18% | 3.9 | 20 |
| 9 | 1482413 | 1478841 | 0.24% | 1.9 | 51 |
| 10 | 1483355 | 1479214 | 0.28% | 6.7 | 55 |
| Averages | | | 0.27% | 3.5 | 38 |

Table 1. MMS Results on ten 2500 variable ORLIB UBQP

It can be difficult to compare solutions and times from other papers due to the lack of detail associated when reporting only the time to best solution. Time to best solution is a single number that does not adequately express the search's performance over time. For example, it does not convey information about time to the previous best solution which may also be a very good solution. To address this we present a set of graphs showing the search histories for these ten problems.

Figure 3 shows the percent of best known value on the y-axis and the time to that value on the x. In these results the y-axis is scaled to better illustrate the changes. If scaled from 0, little observable difference would be seen. Figure 3 shows that the first solution found for all problems was always within about 1% of the best known. Thus, our approach yields solutions within 1% of best known in under one second.

All problems were set to run for 2500 iterations and the average time to finish was consistently 120 seconds. Thus, the largest improvements were found early, generally within about the first 50 iterations. This is not too surprising because as the solution improves, the screening level to allow in more solutions for further improvement also increases. Figure 3 also illustrates that on six of the ten problems, an initial solution is found in about half the time reported because small improvements are found. For example, problem 1 finds a solution within 0.2% of best at 2 seconds, then continues to find small improvements.

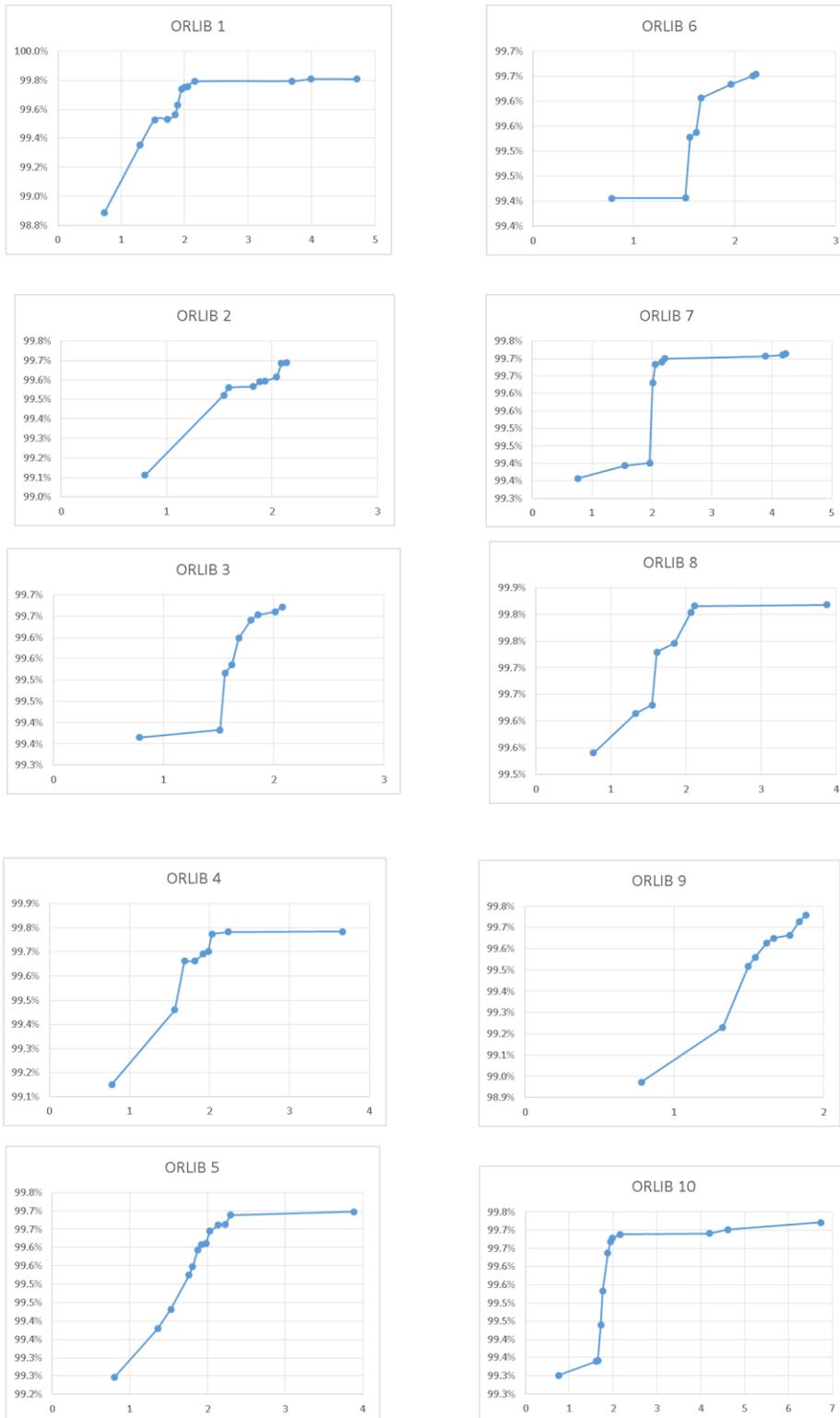

Figure 3. ORLIB UBQP Solution Improvement over Time

Additional tests were performed on the larger, denser UBQP found in (Palubeckis, 2004). These problems have five, six and seven thousand variables and hence are more time-consuming than the 2500 variable ORLIB problems. Our approach also performed well on this problem set, finding a solution within an average of 0.37% of best known in under an average of 53 seconds.

Problem p5000.4 is one outlier from the other times reported, but as illustrated in Figure 4, a solution within 0.8% of best known is found within 10 seconds and then smaller improvements occur over a longer period of time. Reporting search results over time takes up more room and can be difficult to summarize when working with a large set of problems, but reporting a single number to represent the performance of the entire search can be misleading. Using the 0.76% of best known, 10 second result for problem p5000.4 yields an average of 0.37% from best known in an average 26 seconds for the eleven problems.

|           |                  |                 |                              | Times (sec) |                   |
|-----------|------------------|-----------------|------------------------------|-------------|-------------------|
| Problem # | best reported    | MMS Solution    | % diff from best reported    | MMS         | Yang & Glover     |
| p5000.1   | 8559680          | 8538223         | 0.25%                        | 32          | 387               |
| p5000.2   | 10836019         | 10811401        | 0.23%                        | 35          | 609               |
| p5000.3   | 10489137         | 10455640        | 0.32%                        | 21          | 967               |
| p5000.4   | 12252318         | 12185665        | 0.54%                        | 315         | 767               |
| p5000.5   | 12731803         | 12693206        | 0.30%                        | 12          | 726               |
| p6000.1   | 11384976         | 11338236        | 0.41%                        | 15          | 1136              |
| p6000.2   | 14333855         | 14253558        | 0.56%                        | 16          | 1076              |
| p6000.3   | 16132915         | 16086325        | 0.29%                        | 19          | 1053              |
| p7000.1   | 14478676         | 14442479        | 0.25%                        | 34          | 1917              |
| p7000.2   | 18249948         | 18198092        | 0.28%                        | 41          | 1591              |
| p7000.3   | 20446407         | 20354280        | 0.45%                        | 45          | 1503              |
|           |                  | Averages        | 0.35%                        | 53          | 1067              |

Table 2. Results based on Palubeckis' Problem Set

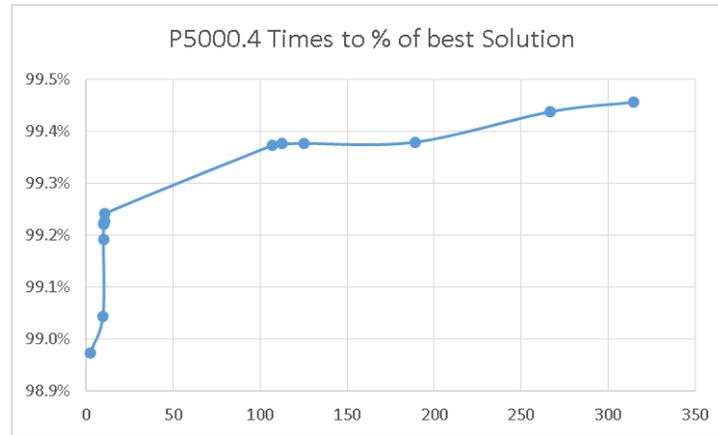

Figure 4. Solution Improvement over time p5000.4

**Conclusions**

The GPU enables rapid evaluation of objective functions involving matrix multiplication. This capability allows a large number of multi-starts to be generated and screened for an initial simple optimization. This paper found that this approach generated high quality solutions very quickly to two sets of benchmark UBQP but in general did not find best known solutions. Thus the approach may place too much emphasis on diversification of the search and not enough on intensification or improvement of the solution.

**References**

Barahona, F., M. Grötschel, M. Jünger & G. Reinelt, 1988. An application of combinatorial optimization to statistical physics and circuit layout design. *Operations Research,* 36(3), pp. 493-513.

Alidaee, B., Glover, F., Kochenberger, G. & Rego, C., 2005. A new modeling and solution approach for the number partitioning problem. *Journal of Applied Mathematics & Decision Sciences,* 2005(2), pp. 113-121.

Alidaee, B., Kochenberger, G. & Ahmadian, A., 1994. 0-1 Quadratic programming approach for optimum solutions of two scheduling problems. *International Journal of Systems Science,* 25(2), pp. 401-408.

Beasley, J. E., 1990. *Welcome to OR-Library.* [Online]
Available at: http://people.brunel.ac.uk/~mastjjb/jeb/info.html
[Accessed June 2015].

Boros, E. & Hammer, P., 1991. The max-cut problem and quadratic 0-1 optimization polyhedral aspects, relaxations and bounds. *Annals of Operations Research,* 33(4127), pp. 151-180.

Glover, F., 1998. A Template for Scatter Search and Path Relinking. Volume 1363, pp. 1-51.

Glover, F., Alidaee, B., Rego, C. & Kochenberger, G., 2002. One-pass heuristics for large-scale unconstrained binary quadratic problems. *European Journal of Operational Research,* 137(2), pp. 272-287.

Kochenberger, G., Glover, F. & Lewis, K., 2005. Using the unconstrained quadratic program to model and solve Max 2-SAT problems. *International Journal of Operational Research,* 1(1), pp. 89-100.

Kochenberger, G. et al., 2013. Solving large scale max cut problems via tabu search. *Journal of Heuristics,* 19(4), pp. 565-571.

nVidia, 2015. *CUDA Toolkit.* [Online]
Available at: https://developer.nvidia.com/cuda-toolkit
[Accessed August 2015].

Palubeckis, G., 2004. Multistart Tabu Search Strategies for the Unconstrained Binary Quadratic Optimization Problem. *Annals of Operations Research,* 131(4127), pp. 259-282.

Pardalos, P. & Xue, J., 1994. The maximum clique problem. *Journal of global Optimization,* 4(3), pp. 301-328.

Wang, Y., Lu, Z., Glover, F. & Hao, J., 2012. Path relinking for unconstrained binary quadratic programming. *European Journal of Operational Research,* 223(3), pp. 595-604.